\pdfoutput=1

\documentclass[11pt]{article}

\usepackage[dvipsnames]{xcolor}

\usepackage[final]{acl}

\usepackage{times}
\usepackage{latexsym}

\usepackage[T1]{fontenc}

\usepackage[utf8]{inputenc}

\usepackage{microtype}

\usepackage{inconsolata}

\usepackage{graphicx}

\usepackage{booktabs}    %
\usepackage{multirow}    %

\usepackage{amsfonts, amssymb, amsmath, soul, bm, paralist}
\usepackage[capitalise]{cleveref}
\usepackage[most]{tcolorbox}

\usepackage{lipsum}

\title{The Medium Is Not the Message: \\ Deconfounding Document Embeddings via Linear Concept Erasure}

\author{
\begin{tabular}{ccc}
\textbf{Yu Fan\textsuperscript{1}} & \textbf{Yang Tian\textsuperscript{2}} & \textbf{Shauli Ravfogel\textsuperscript{3}} \\
\normalfont\texttt{yufan@ethz.ch} & \normalfont\texttt{yang.tian@uzh.ch} & \normalfont\texttt{shauli.ravfogel@gmail.com} \\
\textbf{Mrinmaya Sachan\textsuperscript{1}} & \textbf{Elliott Ash\textsuperscript{1}\thanks{Equal supervision.}} & \textbf{Alexander Hoyle\textsuperscript{1}\footnotemark[1]} \\
\normalfont\texttt{msachan@ethz.ch} & \normalfont\texttt{ashe@ethz.ch} & \normalfont\texttt{hoylea@ai.ethz.ch}
\end{tabular}
\\[1.5ex]
\textsuperscript{1}ETH Zurich \quad
\textsuperscript{2}University of Zurich \quad
\textsuperscript{3}Center for Data Science, New York University
}

\begin{document}
\maketitle
\begin{abstract}%
Embedding-based similarity metrics between text sequences can be influenced not just by the content dimensions we most care about, but can also be biased by spurious attributes like the text's source or language. These document confounders cause problems for many applications, but especially those that need to pool texts from different corpora. This paper shows that a debiasing algorithm that removes information about observed confounders from the encoder representations substantially reduces these biases at a minimal computational cost. Document similarity and clustering metrics improve across every embedding variant and task we evaluate---often dramatically.
Interestingly, performance on out-of-distribution benchmarks is not impacted, indicating that the embeddings are not otherwise degraded.\footnote{Code and data available at \url{https://github.com/y-fn/deconfounding-embeddings}.}
\end{abstract}

\section{Introduction}
\begin{figure}[t]
  \centering
  \includegraphics[width=\columnwidth]{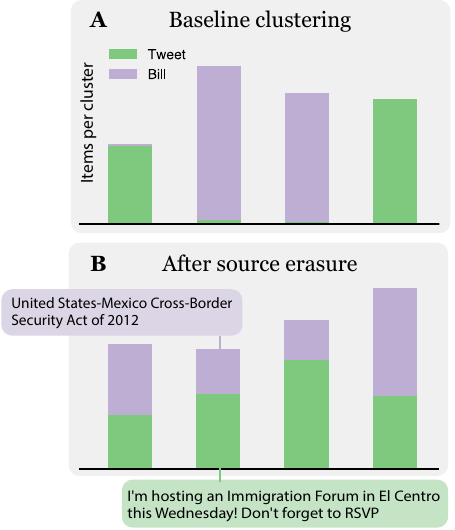}
  \caption{Clustering text embeddings from disparate sources (here, U.S. congressional bill summaries and senators' tweets) can produce clusters where one source dominates (Panel A). Using linear erasure to remove the source information produces more evenly balanced clusters that maintain semantic coherence (Panel B; sampled items relate to immigration). Four random clusters of $k$-means shown ($k$=25), trained on a combined 5,000 samples from each dataset.}
  \label{fig:teaser}
\end{figure}

Suppose a political scientist is studying U.S. political discourse.
They have access to two data sources: Twitter posts from senators and summaries of congressional bills. %
A natural first step in data exploration is to embed the texts (e.g., with a sentence transformer; \citealt{reimers-gurevych-2019-sentence}) and then cluster them (e.g., with $k$-means).
However, some clusters will predominantly contain items from one source or the other, because systematic differences between sources dominate the distances that $k$-means relies on (\cref{fig:teaser}A). 

Text embeddings, generated by pretrained models, capture a wide range of information about text, including topical, semantic, stylistic, multilingual, and syntactic features.
These models are typically trained with the goal of ``making semantically similar sentences close in vector space'' \cite{reimers-gurevych-2019-sentence}.
However, this objective can cause spurious correlations---such as between domain and topic---to be encoded as unintended relationships. As \citet{thompson-mimno-2018-authorless} observe: ``collections are often constructed by combining documents from multiple sources, [so the] most prominent patterns in a collection simply repeat the known structure of the corpus.''\footnote{Their analysis focuses on bag-of-words topic models rather than text embeddings, so their vocabulary-based approach does not translate to our setting.}
It therefore would seem useful to remove unwanted information from text representations.

Indeed, adjusting embeddings to remove confounding information is exactly what we do in this work.
Adapting the algorithm from \citet{belrose2023leace} for linear concept erasure, we remove embedding subspaces that are predictive of the confounding variables, which can bias measures of document distance. In the above example from U.S. politics, we residualize out the source information (Twitter or bills), producing adjusted embeddings for which similarity metrics load on the semantic content rather than the source (\cref{fig:teaser}B). As another practical example, in a multilingual corpus, we residualize out the subspace that is predictive of language, leading to document distance metrics that are driven by content, rather than language.

Extensive tests show that the adjusted embeddings perform significantly better for clustering and similarity search. For example, in a multilingual document search setting, Recall@1 increases from 0.18 to 0.83. Importantly (and surprisingly), there is also no reduction in performance when using the adjusted embeddings on unseen datasets and tasks from a standard retrieval benchmark \cite{muennighoff-etal-2023-mteb, enevoldsen-etal-2024-dansk}, suggesting erasure does not harm embedding quality.

The approach is computationally inexpensive, involving only linear transformations on pretrained embeddings. %
To support practical use, we release a wrapper that streamlines encoding, fitting the deconfounder, and adjusting embeddings. In addition, we provide several evaluation datasets labeled both with confounders (language and source) and semantic content (e.g., topic).
In sum, we:
\begin{compactitem}
    \item Formally demonstrate how erasure removes confounding information from document similarities (\S\ref{sec:background});
    \item Construct a benchmark of paired data to measure the impact of confounding attributes on embedding performance (\S\ref{sec:setup});
    \item Evaluate a diverse set of embedding methods,  showing that observable features such a text’s source can reduce the utility of text embeddings in applied settings (\S\ref{sec:results});
    \item Show that applying a linear erasure algorithm to remove observed confounders can effectively mitigate such issues---sometimes dramatically---without degrading other aspects of performance (\S\ref{sec:mteb}).
\end{compactitem}

\begin{figure}[ht]
    \centering
    \includegraphics[width=0.9\linewidth]{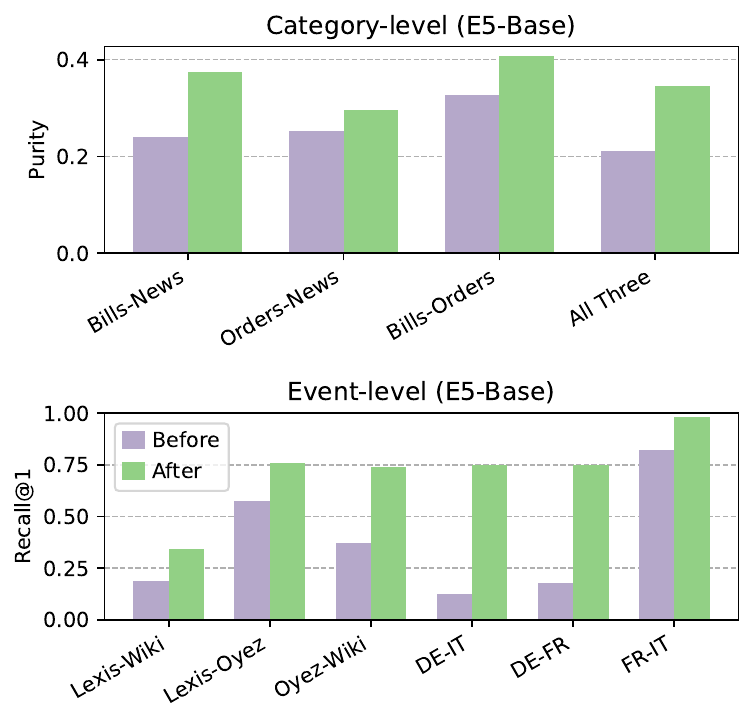}
    \caption{Performance of the multilingual E5-base model before and after erasure. Top: category-level results from the Comparative Agendas Project (metric: purity). Bottom: event-level results from SCOTUS and multilingual Swiss case summaries (metric: recall@1). Detailed results for other models and datasets in \cref{tab:model-scores-category-nested,tab:event-scores-swiss,tab:event-scores-scotus,tab:model-scores-pert-semeval}. }
    \label{fig:e5-large}
\end{figure}

\begin{figure}[t]
\centering
\includegraphics[width=\columnwidth]{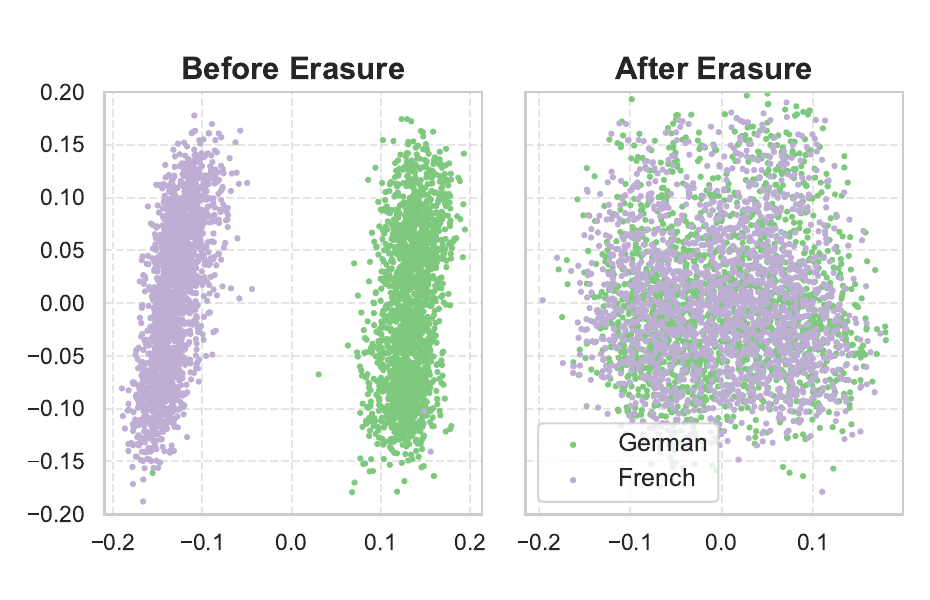}
\caption{PCA projection of text embeddings before and after LEACE. Data are paired Swiss court case summaries in German (green) and French (purple). We deploy multilingual E5 as the embedding model. The first principal component recovers the two languages almost exactly.}
\label{fig:erasure-illustration}
\end{figure}

\section{Background}\label{sec:background}

Many downstream tasks such as nearest-neighbor search, clustering, retrieval, topic discovery etc. rely on assessing how ``close'' two documents are in an embedding space. An effective distance metric should rank pairs by semantic relatedness rather than by superficial attributes like author, language, or publication venue. In practice, however, pretrained embedding models often encode these incidental signals, since they occur frequently during training and help optimize self-supervised objectives. When such signals correlate with content, distance measures become biased and can undermine empirical conclusions.

\paragraph{Embedding text sequences}

Sentence-level embeddings position semantically similar documents close to each other in a vector space \citep{kiros2015skip, conneau-etal-2017-supervised, cer-etal-2018-universal, reimers-gurevych-2019-sentence}. Modern systems typically begin with a transformer encoder pretrained on masked-language modeling, then refine it on hundreds of millions of contrastive pairs drawn from diverse corpora \citep{reimers-gurevych-2019-sentence}. This approach underpins state-of-the-art performance in retrieval \citep{asai-etal-2021-xor, thakur2021beir, zhang-etal-2023-miracl}, clustering \citep{aggarwal2012survey}, and classification \citep{maas-etal-2011-learning}.

Contrastive batches are often drawn from a single source, enabling the model to focus on internal semantics \citep{nussbaum2024nomicembedtrainingreproducible}. A side effect is that different sources may occupy distinct regions of the embedding space, especially when cross-source positives are scarce. Multilingual models exhibit a similar problem: even when trained with translation pairs \citep{wang2024multilingual}, large amounts of monolingual data tends to push languages apart.

Despite attempts to construct comparable contrastive pairs, the resulting embeddings still encode confounding information. Platform-specific jargon and style can dominate representations. In addition, language itself may serve as a proxy for topic or geography. For authors and outlets, stylistic markers linked to gender or ideology can act as shortcuts for similarity. Because these attributes correlate with content, they function as \emph{observed confounders} in distance-based analyses.

\paragraph{The document comparison problem.}
More formally, let $\bm{X} \in \mathbb{R}^d$ denote the embedding of a random document. For a particular document $d_i$, we write its realization as $\bm{x}_i$ and assume $\lVert \bm{x}_i \rVert = 1$. In general, we use bold uppercase letters (e.g., $\bm{X}$) for random variables and lowercase letters ($\bm{x}_i$) for their realizations. Erased variables use a tilde ($\tilde{\mathrm{B}}$). Assume a linear decomposition for the embedding:  
\begin{equation}
\bm{X} \;=\; \mathrm{B}_z \bm{Z} \;+\; \mathrm{B}_c \bm{C} \;+\; \mathrm{B}_u
\bm{U} \;+\; \bm{\mathcal{E}},
\label{eq:decomp}
\end{equation}
where $\bm{Z}$ is a random variable representing the \emph{semantic content} of interest (e.g.\ topic); $\bm{C}$ the \emph{observed confounders} (source, language, author traits); and $\bm{U}$ the \emph{unobserved confounders}. $\mathrm{B}_z$,$\mathrm{B}_c$, $\mathrm{B}_u$ are loading matrices. $\bm{Z}$, $\bm{C}$, $\bm{U}$ and the noise $\bm{\mathcal{E}}$ are zero-mean and uncorrelated with one another.

Similarity is measured with the dot product, where $x_0$ and $x_1$ are two IID draws of $\bm{X}$:  
\begin{equation}
y_{01}\;=\;\bm{x}_0^{\!\top}\bm{x}_1.
\label{eq:sim_raw}
\end{equation}
Under the model expressed in \eqref{eq:decomp}, expanding this dot product creates multiple entangled factors:
\begin{align}
y_{01}
  \;=\; \bm{z}_0^{\!\top}\Gamma_{zz}\bm{z}_1
       \;+\; \bm{z}_0^{\!\top}\Gamma_{zc}\bm{c}_1\\
       \;+\; \bm{z}_0^{\!\top}\Gamma_{zu}\bm{u}_1
       \;+\; \bm{c}_0^{\!\top}\Gamma_{zc}\bm{z}_1
       \;+\; \ldots,
\label{eq:sim_parts}
\end{align}
where $\Gamma_{jk}$ \;=\; $B_j^{\!\top}B_k$.
Only the first term reflects the semantic proximity we care about; the others bias any analysis based on $y_{01}$.

\paragraph{Debiasing and concept erasure.}

Concept erasure techniques aim to remove a targeted feature (e.g.\ gender) from an embedding. This is typically achieved by projecting out the corresponding subspace, thereby reducing bias and enabling analysis of model behavior without that feature \citep{Ravfogel2022LinearAC, belrose2023leace}.

Early debiasing work on word vectors identified a ``bias direction'' (e.g.\ race) and removed its projection \citep{Bolukbasi2016ManITA}. Later studies showed that the removed signal remained recoverable \citep{gonen-goldberg-2019-lipstick}, motivating stronger linear methods such as Iterative Null-space Projection \citep[INLP,][]{ravfogel-etal-2020-null}, Linear Adversarial Concept Erasure \citep[LACE,][]{Ravfogel2022LinearAC}, and LEAst-squares Concept Erasure \citep[LEACE,][]{belrose2023leace}. These approaches seek an affine transformation that eliminate all linear correlation with the protected attribute while altering the representations as little as possible.

An important special case of these kinds of concept erasure is \emph{linear concept erasure}, where the goal is to prevent linear adversaries from predicting the information we aim to remove. This is usually achieved in the form of a projection matrix that neutralizes a subspace that is associated with the concept $\bm{C}$. Following \citet{Ravfogel2022LinearAC},  \citet{belrose2023leace} derived sufficient and necessary conditions for achieving \emph{linear guardedness} \citep{ravfogel-etal-2023-linear}, a situation where \emph{no linear classifier} can recover the concept $\bm{C}$ and achieve a loss lower than that of a trivial predictor that always predicts the majority class. Specifically, they derive a \emph{linear projection matrix} $\bm{P}^\ast$  such that:

\begin{align}
\mathrm{P}^\ast &= \arg \min_{\mathrm{P} \in \mathbb{R}^{d \times d}} \mathbb{E} \left [ \| \mathrm{P} \bm{X} - \bm{X} \|^2 \right] \\
    & \text{subject to Cov}(\mathrm{P}\bm{X}, \bm{C}) = 0.
\end{align}

The covariance constraint ensures the erasure of linear information, while the first objective minimizes \emph{distortion} of the representation space. It turns out that this objective has a closed-form solution in the form of 

\begin{equation}
\mathrm{P}^\ast = \mathrm{I} \;-\; \mathrm{W}^{\dagger}\!(\mathrm{W}\mathrm{\Sigma}_{XC})(\mathrm{W}\mathrm{\Sigma}_{XC})^{\dagger}\mathrm{W}
\label{eq:leace_map}
\end{equation}
where $\mathrm{W}=\mathrm{\Sigma}_{XX}^{-1/2}$ is a whitening matrix, $\mathrm{W}^\dagger$ the pseudoinverse of $\mathrm{W}$, and
\[
\mathrm{\Sigma}_{XC}=\operatorname{Cov}(\bm{X}, \bm{C}),
\mathrm{\Sigma}_{XX}=\operatorname{Cov}(\bm{X}),
\bm{\mu}=\mathbb{E}[\bm{X}].
\]

This condition is proved to be sufficient and necessary for achieving \emph{linear guardedness}, i.e., the inability of any linear classifier to recover the attribute $\bm{C}$ from the embeddings.
In other words, Eq. \eqref{eq:leace_map} together with $\bm{b}=\bm{\mu}-\mathrm{P}\bm{\mu}$ defines the unique affine map that removes \emph{all} linear correlations with the observed confounder $\bm{C}$ while modifying the embeddings as little as possible. 

For any document, the \emph{adjusted} embedding is $\tilde{\bm{x}}_i = \mathrm{P} \bm{x}_i + \bm{b}$. Applying this LEACE map to the realization of the structural decomposition in~\eqref{eq:decomp}:
\begin{align}
\tilde{\bm{x}}_i
  &= P\bigl(\mathrm{B}_z \bm{z}_i + \mathrm{B}_c \bm{c}_i + \mathrm{B}_u \bm{u}_i + \bm{\varepsilon}_i\bigr) + \bm{b}
     \notag\\
  &= \tilde{\mathrm{B}}_z \bm{z}_i + \tilde{\mathrm{B}}_u \bm{u}_i + \mathrm{P}\bm{\varepsilon}_i,
\label{eq:PX_expand}
\end{align} 
where $\tilde{\mathrm{B}}_j=\mathrm{P}\mathrm{B}_j$. Note that the middle term vanishes, following from the constraint $\operatorname{Cov}(\tilde{\bm{X}}, \bm{C}) = 0$, which ensures that the space spanned by $\mathrm{B}_c$ is removed. %
In turn, the estimand for the document similarity,
\begin{align}
\tilde{y}_{01}
  \;=\;\tilde{\bm{x}}_0^{\!\top}\tilde{\bm{x}}_1
  \;=\; \bm{z}_0^{\!\top}\tilde{\mathrm{\Gamma}}_{zz} \bm{z}_1
  \;+\; \bm{z}_0^{\!\top}\tilde{\mathrm{\Gamma}}_{zu} \bm{u}_1\nonumber\\
  \;+\; \bm{u}_0^{\!\top}\tilde{\mathrm{\Gamma}}_{uz} \bm{z}_1
       \;+\; \bm{u}_0^{\!\top}\tilde{\mathrm{\Gamma}}_{uu} \bm{u}_1 + P\bm{\varepsilon}_i,
\label{eq:sim_adjusted}
\end{align}

is also purged of $\bm{C}$ (furthermore, in expectation it does not include the cross terms, as they are uncorrelated under our assumptions). Note, however, that the projection alters the geometry of the remaining components. $\tilde{\bm{x}}$ is now based on $\mathrm{P}\mathrm{B}_z\bm{z}$, which may not be equal to $\mathrm{B}_z\bm{z}$, depending on the intensity and nature of the dependence between $\bm{z}$ and $\bm{C}$. So the LEACE algorithm might also add bias to similarity metrics through its adjustment of this term. Further, the (adjusted) unobserved confounder $\bm{u}$ remains, and it is unclear how the deconfounding by LEACE would either increase or reduce bias from $\bm{u}$.

\begin{table}[ht]\label{tab:cat-data}
\centering\small
\begin{tabular}{lc c}
\midrule
\emph{Category-level Data} & $N_\text{total}$ & Categories  \\
\cmidrule{1-3}
\multicolumn{3}{l}{\ \textbf{CAP Data}} \\
\quad Bills -- Orders          & 1,902 & 21 \\
\quad Bills -- Newspapers      & 2,613 & 21 \\
\quad Orders -- Newspapers     & 1,907 & 21 \\
\quad All Three Sources       & 3,211 & 21 \\
\\
\midrule
\emph{Event-level Data} & $N_\text{paired}$& $N_\text{unpaired}$ \\
\midrule
\multicolumn{3}{l}{\ \textbf{SCOTUS Cases}} \\
\quad Wikipedia -- LexisNexis    & 2,048 & 1,518 \\
\quad Wikipedia -- Oyez          & 1,560 & 1,762 \\
\quad LexisNexis -- Oyez         & 2,048 & 2,075 \\
\\
\multicolumn{3}{l}{\ \textbf{SemEval News Articles}} \\
\quad EN -- Non-EN       & 888 & 0 \\
\\
\multicolumn{3}{l}{\ \textbf{Swiss Court Cases}} \\
\quad DE -- FR           & 2,048 & 1,760 \\
\quad DE -- IT           & 2,048 & 1,760 \\
\quad FR -- IT           & 2,048 & 1,760 \\
\bottomrule
\end{tabular}
\caption{Dataset statistics. The data cover a variety of domains and languages.}
\label{tab:event-data}
\end{table}

\section{Experimental Setup}\label{sec:setup}

Our evaluation settings are designed to approximate real-world use cases and involve datasets from multiple corpora.
They are divided into two groups, \emph{category-level} and \emph{event-level} data, both aiming to measure the same thing: the extent to which documents that share a common label have similar embeddings.

The approach is the same across all datasets: create a vector of concept labels $\bm{c}$ to erase, using known metadata (here, a text's source or language).
Then, pass each text item through the embedding model to obtain a matrix $\mathbf{X}$.
Fit LEACE on $(\mathbf{X},\bm{c})$ to learn the whitening and projection matrices, then apply the transformations back to $\mathbf{\tilde{X}}$.\footnote{For the out-of-sample experiments in \cref{sec:mteb}, the transformations are applied to novel benchmark data $\mathbf{X}'$.}

\subsection{Category-level Data}\label{sec:setup:category}

Recalling the motivating example from the introduction, imagine a researcher clusters documents from different sources (like news articles and court cases), with the hope that each cluster contains documents that fall under a coherent topic.

We measure progress on this task by relying on a common set of ground-truth category labels, like ``Education'', that cover multiple datasets.
The goal is that the assigned clusters align with the categories, even if the constituent documents come from different sources.

\paragraph{Datasets.} We use datasets from the Comparative Agendas Project (CAP), which provides a coding framework for analyzing policy activities across time and between countries \cite{jones2023policy}.

We use texts from three sources: newspaper articles\footnote{\url{https://comparativeagendas.net/project/pennsylvania}}, congressional bill summaries (\citealt{wilkerson2023policy}, taken from \citealt{hoyle-etal-2022-neural}), and executive orders \cite{jones2023executive}. We evaluate each pair of sources separately, as well as all three simultaneously.

\paragraph{Metrics and Methodology.} We measure alignment between ground-truth category labels and assigned clusters with two metrics. Following \citet{poursabzi-sangdeh-etal-2016-alto}, we use purity, which quantifies to what extent each cluster contains items from a single gold category, and the Adjusted Rand Index, a chance-corrected metric that measures the similarity of two clusterings.

The erased concept is the \emph{source} for each of the four settings (\cref{tab:event-data}).
When generating clusters, we follow a standard practice and apply $k$-means to the text embeddings for each document \cite{zhang-etal-2022-neural}.\footnote{We set $k$ = 21, the total number of categories in the data. Improvements are robust to different $k$, see~\cref{fig:purity} in appendix.}

\begin{table*}[t]
\centering
\resizebox{\textwidth}{!}{%
\begin{tabular}{l
    *{2}{cc}  %
    *{2}{cc}  %
    *{2}{cc}  %
    *{2}{cc}  %
}
\toprule
\multirow{3}{*}{Model}
& \multicolumn{4}{c}{Bills \& News}
& \multicolumn{4}{c}{Orders \& News}
& \multicolumn{4}{c}{Bills \& Orders}
& \multicolumn{4}{c}{All Three Sources} \\
\cmidrule(lr){2-5} \cmidrule(lr){6-9} \cmidrule(lr){10-13} \cmidrule(lr){14-17}
& \multicolumn{2}{c}{\texttt{Purity}} & \multicolumn{2}{c}{\texttt{ARI}}
& \multicolumn{2}{c}{\texttt{Purity}} & \multicolumn{2}{c}{\texttt{ARI}}
& \multicolumn{2}{c}{\texttt{Purity}} & \multicolumn{2}{c}{\texttt{ARI}}
& \multicolumn{2}{c}{\texttt{Purity}} & \multicolumn{2}{c}{\texttt{ARI}} \\
\cmidrule(lr){2-3} \cmidrule(lr){4-5}
\cmidrule(lr){6-7} \cmidrule(lr){8-9}
\cmidrule(lr){10-11} \cmidrule(lr){12-13}
\cmidrule(lr){14-15} \cmidrule(lr){16-17}
& Before & After & Before & After
& Before & After & Before & After
& Before & After & Before & After
& Before & After & Before & After \\
\midrule
MiniLM     & 0.346 & \textbf{0.507} & 0.148 & \textbf{0.268} & 0.329 & \textbf{0.463} & 0.123 & \textbf{0.228} & 0.391 & \textbf{0.448} & 0.169 & \textbf{0.226} & 0.269 & \textbf{0.411} & 0.096 & \textbf{0.205} \\
GIST-small & \underline{0.380} & \underline{\textbf{0.549}} & \underline{0.171} & \underline{\textbf{0.328}} & \underline{0.421} & \textbf{0.515} & \underline{0.200} & \textbf{0.283} & 0.422 & \underline{\textbf{0.513}} & 0.191 & \underline{\textbf{0.275}} & 0.330 & \underline{\textbf{0.483}} & \underline{0.131} & \underline{\textbf{0.259}} \\
E5-small   & 0.260 & \textbf{0.414} & 0.085 & \textbf{0.207} & 0.289 & \textbf{0.290} & 0.099 & \textbf{0.101} & 0.319 & \textbf{0.422} & 0.123 & \textbf{0.190} & 0.237 & \textbf{0.356} & 0.069 & \textbf{0.166} \\
\\
MPNet      & 0.365 & \textbf{0.504} & 0.162 & \textbf{0.282} & 0.377 & \textbf{0.444} & 0.151 & \textbf{0.217} & \underline{0.461} & \textbf{0.493} & \underline{0.229} & \textbf{0.256} & \underline{0.334} & \textbf{0.481} & 0.130 & \underline{\textbf{0.259}} \\
GIST-base  & 0.373 & \textbf{0.534} & 0.157 & \textbf{0.312} & 0.380 & \underline{\textbf{0.534}} & 0.165 & \underline{\textbf{0.309}} & 0.425 & \textbf{0.498} & 0.188 & \textbf{0.262} & 0.320 & \textbf{0.470} & 0.054 & \textbf{0.147} \\
E5-base    & 0.240 & \textbf{0.375} & 0.072 & \textbf{0.175} & 0.252 & \textbf{0.297} & 0.075 & \textbf{0.108} & 0.328 & \textbf{0.407} & 0.130 & \textbf{0.173} & 0.212 & \textbf{0.346} & 0.130 & \textbf{0.173} \\
Nomic-v2   & 0.324 & \textbf{0.463} & 0.122 & \textbf{0.250} & 0.331 & \textbf{0.353} & 0.127 & \textbf{0.161} & 0.386 & \textbf{0.442} & 0.159 & \textbf{0.218} & 0.249 & \textbf{0.411} & 0.073 & \textbf{0.196} \\
\\
MXB-large  & 0.328 & \textbf{0.493} & 0.134 & \textbf{0.279} & 0.332 & \textbf{0.524} & 0.127 & \textbf{0.281} & 0.420 & \textbf{0.487} & 0.188 & \textbf{0.263} & 0.299 & \textbf{0.410} & 0.112 & \textbf{0.199} \\
GIST-large & 0.361 & \textbf{0.492} & 0.148 & \textbf{0.295} & 0.375 & \textbf{0.471} & 0.153 & \textbf{0.258} & 0.418 & \textbf{0.495} & 0.195 & \textbf{0.258} & 0.294 & \textbf{0.434} & 0.106 & \textbf{0.226} \\
E5-large   & 0.224 & \textbf{0.373} & 0.066 & \textbf{0.170} & 0.273 & \textbf{0.283} & 0.082 & \textbf{0.103} & 0.327 & \textbf{0.366} & 0.104 & \textbf{0.152} & 0.211 & \textbf{0.297} & 0.055 & \textbf{0.124} \\
\bottomrule
\end{tabular}
}
\caption{
Cluster alignment metrics on the ``category-level'' Comparative Agendas Project datasets (\S\ref{sec:setup:category}), before and after linear concept erasure. Here, the erased concept is the \emph{source} (top row). We set $k$ = 21, the total number of categories in the CAP datasets. Erasure substantially improves cluster alignment for every combination of sources across all embedding models. Bolded scores indicate performance improvements after erasure; underlined scores mark the highest value in each column.}
\label{tab:model-scores-category-nested}
\end{table*}

\subsection{Event-level Data}\label{sec:setup:event}
Now imagine that a practitioner wants to understand how a common event---a court case, a natural disaster---is portrayed by distinct sources or languages.
If they have access to one document discussing the event, how can they best find others?

\paragraph{Datasets.} We rely on three paired datasets, which link documents depicting the same event in different sources or languages.

\textbf{Super-SCOTUS} \cite{fang-etal-2023-super} contains case summaries from the U.S. Supreme Court sourced from LexisNexis and Oyez. In addition, we scrape case summaries from Wikipedia. This results in 1,518 pairs of LexisNexis and Wikipedia case summaries, 2,075 from LexisNexis and Oyez, and 780 pairs from Wikpedia and Oyez.

\textbf{SemEval 2022 Task 8} \cite{chen-etal-2022-semeval} assesses the similarity between pairs of multilingual news articles. We obtain 444 pairs of news articles that depict similar events in different languages, namely English and non-English (Spanish, German, and Chinese).

A third dataset is derived from \textbf{SwilTra-Bench} \cite{niklaus2025swiltra}, which contains parallel summaries of leading Swiss court decisions from the Federal Supreme Court of Switzerland in German, French, and Italian.

\paragraph{Methodology and Metrics.} To accurately simulate real-world conditions, in which only partially paired data is available and the remaining data is unpaired and derived from different sources, we retain up to 1,024 data pairs for each applicable setting. We treat the remainder of the data as unpaired by randomly discarding one example from each pair. Thus, data is considered unpaired either because paired data was unavailable from the original sources or because one item from a pair was randomly removed. In each setting, we pool together the paired and unpaired data and subsequently use this combined dataset to train the LEACE eraser, aiming to remove source-specific information.

We evaluate whether each paired item can retrieve its counterpart from the pooled dataset using \textbf{Recall@1 and @10}, the proportion of correct matches that appear in the top $k$ retrieved results.

\subsection{Embedding Models}
Our experiments use ten embedding models of varying sizes and dimensionality (appendix \cref{tab:embed}). This set includes multilingual and monolingual variants, as well as models with instruction fine-tuning: \texttt{MiniLM\footnote{\url{https://huggingface.co/sentence-transformers/all-MiniLM-L6-v2}}}, \texttt{GIST-small}, \texttt{GIST-base}, \texttt{GIST-large} \cite{solatorio2024gistembed}, multilingual \texttt{E5-small}, \texttt{E5-base}, \texttt{E5-large} \cite{wang2024multilingual}, \texttt{all-mpnet-base-v2} \cite[][]{song2020mpnet}, \texttt{Nomic-v2} \cite{nussbaum2025trainingsparsemixtureexperts}, and \texttt{MXB-large} \cite{li2023angle, lee2024open}.\looseness=-1

\section{Primary Results}\label{sec:results}
We first discuss the results on the category-level datasets, then turn to the event-level.
In brief, erasure improves embeddings across the board---over all models, metrics, and datasets (when performance is not already saturated). A summary of results for a single model is in Fig.~\ref{fig:e5-large}.

\subsection{Category-level}

In all four source pairings from the CAP dataset, erasing source-specific information with LEACE consistently improves clustering quality (\cref{tab:model-scores-category-nested}).
In the \textit{Bills–Newspapers} comparison, all ten models show marked improvements, with gains in ARI ranging from +0.104 (E5-large) to +0.157 (GIST-small), and purity increases as high as +0.169 (GIST-small). Although the magnitude of improvement varies, this pattern persists in the \textit{Orders–Newspapers} comparison. While most models benefit substantially, multilingual models such as E5-small and E5-large show only marginal gains, suggesting that source signal may be less distinct in this pairing.

The \textit{Bills–Orders} setting yields more moderate improvements, yet the gains remain consistent across model scales. Finally, the \textit{All Three Sources} setting demonstrates that LEACE generalizes to more complex source distributions. Smaller-sized models, such as MiniLM and GIST-small, gain over +0.130 in purity and +0.100 in ARI. Even larger models such as GIST-large and MXB-large improve substantially after concept erasure.

Overall, these results demonstrate the robustness of LEACE across diverse source combinations and embedding models, confirming its ability to reduce spurious relationships between items while preserving task-relevant semantic structure.

\begin{table*}[ht]
\centering
\small
\resizebox{\textwidth}{!}{%
\begin{tabular}{l
    *{2}{cc}  %
    *{2}{cc}  %
    *{2}{cc}  %
}
\toprule
\multirow{3}{*}{Model}
& \multicolumn{4}{c}{LexisNexis \& Wikipedia}
& \multicolumn{4}{c}{LexisNexis \& Oyez}
& \multicolumn{4}{c}{Oyez \& Wikipedia} \\
\cmidrule(lr){2-5} \cmidrule(lr){6-9} \cmidrule(lr){10-13}
& \multicolumn{2}{c}{\texttt{Recall@10}} & \multicolumn{2}{c}{\texttt{Recall@1}}
& \multicolumn{2}{c}{\texttt{Recall@10}} & \multicolumn{2}{c}{\texttt{Recall@1}}
& \multicolumn{2}{c}{\texttt{Recall@10}} & \multicolumn{2}{c}{\texttt{Recall@1}} \\
\cmidrule(lr){2-3} \cmidrule(lr){4-5}
\cmidrule(lr){6-7} \cmidrule(lr){8-9}
\cmidrule(lr){10-11} \cmidrule(lr){12-13}
& Before & After & Before & After
& Before & After & Before & After
& Before & After & Before & After \\
\midrule
MiniLM     & 0.487 & \textbf{0.606} & 0.231 & \textbf{0.313} & 0.890 & \textbf{0.899} & 0.651 & \textbf{0.693} & 0.850 & \textbf{0.924} & 0.623 & \textbf{0.747} \\
GIST-small & 0.563 & \textbf{0.656} & 0.261 & \textbf{0.325} & 0.918 & \textbf{0.943} & 0.702 & \textbf{0.778} & 0.762 & \textbf{0.844} & 0.478 & \textbf{0.599} \\
E5-small   & 0.421 & \textbf{0.673} & 0.176 & \textbf{0.353} & 0.830 & \textbf{0.939} & 0.563 & \textbf{0.789} & 0.689 & \textbf{0.951} & 0.398 & \textbf{0.752} \\
\\
MPNet      & 0.566 & \textbf{0.666} & 0.259 & \textbf{0.337} & 0.926 & \textbf{0.943} & 0.724 & \textbf{0.775} & 0.856 & \textbf{0.911} & 0.565 & \textbf{0.678} \\
GIST-base  & 0.646 & \textbf{0.757} & \underline{0.308} & \textbf{0.412} & 0.939 & \textbf{0.963} & 0.727 & \textbf{0.819} & 0.880 & \textbf{0.950} & 0.628 & \textbf{0.773} \\
E5-base    & 0.414 & \textbf{0.660} & 0.188 & \textbf{0.341} & 0.830 & \textbf{0.940} & 0.575 & \textbf{0.758} & 0.650 & \textbf{0.942} & 0.371 & \textbf{0.737} \\
Nomic-v2   & 0.530 & \textbf{0.701} & 0.254 & \textbf{0.384} & 0.950 & \textbf{0.966} & 0.770 & \textbf{0.820} & 0.903 & \underline{\textbf{0.978}} & 0.658 & \textbf{0.819} \\
\\
MXB-large  & 0.537 & \textbf{0.703} & 0.249 & \textbf{0.376} & 0.928 & \textbf{0.958} & 0.720 & \textbf{0.805} & 0.883 & \textbf{0.960} & 0.654 & \textbf{0.819} \\
GIST-large & \underline{0.657} & \underline{\textbf{0.770}} & 0.305 & \underline{\textbf{0.414}} & \underline{0.954} & \underline{\textbf{0.967}} & \underline{0.787} & \underline{\textbf{0.834}} & \underline{0.947} & \textbf{0.971} & \underline{0.760} & \underline{\textbf{0.826}} \\
E5-large   & 0.479 & \textbf{0.720} & 0.209 & \textbf{0.381} & 0.864 & \textbf{0.949} & 0.636 & \textbf{0.791} & 0.765 & \textbf{0.964} & 0.489 & \textbf{0.792} \\
\bottomrule
\end{tabular}
}
\caption{Document similarity search results on paired ``event-level'' U.S. Supreme Court Summaries (\ref{sec:setup:event}), before and after linear concept erasure. Here, the erased concept is the document's \emph{source}. Erasure improves recall for every setting and model.}
\label{tab:event-scores-scotus}
\end{table*}

\subsection{Event-Level}
At the event level, we present the results with Recall@10 and Recall@1, because only one document is deemed relevant for each query.

\paragraph{U.S. Supreme Court Case Summaries} Applying LEACE consistently improves retrieval performance on the SCOTUS summary data  (\cref{tab:event-scores-scotus}). In both \emph{Wikipedia} pairings, improvements are large and especially pronounced for the E5 family.
For instance, on \emph{LexisNexis-Wikipedia}, E5-small gains +0.177 in Recall@1 and E5-base +0.153.%

Performance before erasure on \textit{LexisNexis--Oyez} is already high, likely because the two have more stylistic elements in common---both being technical summaries based on the original court opinion.
Nonetheless, we still observe more modest but consistent gains. E5-small and E5-base increase Recall@1 by +0.226 and +0.183, respectively, although GIST-base and MXB-large exhibit improvements of only about +0.08.%

Overall, LEACE not only improves representation consistency across heterogeneous legal sources, but also enhances alignment even when initial model performance is already strong.

\paragraph{Swiss Federal Supreme Court Case Summaries} Turning now to multilingual data, we observe that LEACE can be extremely effective, even with already-multilingual embeddings (\cref{tab:event-scores-swiss}).

For all settings on the Swiss court case summary data, nearly every model sees higher recall after applying LEACE. 
The improvements tend to be largest with different language families: German-Italian and German-French.
On \emph{DE-IT}, gains in Recall@1 can reach +0.651 (E5-large); on \emph{DE-FR}, +0.570 (E5-base).
As French and Italian are closer, baseline retrieval is already strong, with some models already having near-perfect Recall@10. This reflects the tendency of related languages to lie closer in embedding space, as shown in prior work on genealogical structure \cite{ostling-kurfali-2023-language} and cross-lingual language representations \cite{sharoff2020finding}.
Still, increases in metrics abound, primarily in the smaller models like MiniLM.
Taken together, LEACE removes source-specific signals even in complex multilingual legal domains.

\paragraph{SemEval News Articles} To avoid bludgeoning the reader with positive results, we briefly outline the results on our other multilingual dataset: all ten models again benefit from erasure (\cref{tab:model-scores-pert-semeval} in the appendix).

\begin{table*}[ht]
\centering
\small
\resizebox{\textwidth}{!}{%
\begin{tabular}{l
    *{2}{cc}  %
    *{2}{cc}  %
    *{2}{cc}  %
}
\toprule
\multirow{3}{*}{Model}
& \multicolumn{4}{c}{DE \& IT}
& \multicolumn{4}{c}{DE \& FR}
& \multicolumn{4}{c}{FR \& IT} \\
\cmidrule(lr){2-5} \cmidrule(lr){6-9} \cmidrule(lr){10-13}
& \multicolumn{2}{c}{\texttt{Recall@10}} & \multicolumn{2}{c}{\texttt{Recall@1}}
& \multicolumn{2}{c}{\texttt{Recall@10}} & \multicolumn{2}{c}{\texttt{Recall@1}}
& \multicolumn{2}{c}{\texttt{Recall@10}} & \multicolumn{2}{c}{\texttt{Recall@1}} \\
\cmidrule(lr){2-3} \cmidrule(lr){4-5}
\cmidrule(lr){6-7} \cmidrule(lr){8-9}
\cmidrule(lr){10-11} \cmidrule(lr){12-13}
& Before & After & Before & After
& Before & After & Before & After
& Before & After & Before & After \\
\midrule
MiniLM     & 0.009 & \textbf{0.086} & 0.003 & \textbf{0.023} & 0.026 & \textbf{0.102} & 0.008 & \textbf{0.030} & 0.146 & \textbf{0.545} & 0.040 & \textbf{0.260} \\
GIST-small & 0.020 & \textbf{0.211} & 0.004 & \textbf{0.063} & 0.041 & \textbf{0.246} & 0.011 & \textbf{0.075} & 0.315 & \textbf{0.771} & 0.101 & \textbf{0.461} \\
E5-small   & 0.093 & \textbf{0.930} & 0.027 & \textbf{0.543} & 0.167 & \textbf{0.937} & 0.051 & \textbf{0.563} & 0.853 & \underline{\textbf{1.000}} & 0.455 & \textbf{0.968} \\
\\
MPNet      & 0.016 & \textbf{0.149} & 0.006 & \textbf{0.048} & 0.053 & \textbf{0.155} & 0.021 & \textbf{0.050} & 0.157 & \textbf{0.646} & 0.052 & \textbf{0.346} \\
GIST-base  & 0.034 & \textbf{0.296} & 0.008 & \textbf{0.092} & 0.076 & \textbf{0.378} & 0.024 & \textbf{0.142} & 0.440 & \textbf{0.873} & 0.167 & \textbf{0.565} \\
E5-base    & 0.380 & \textbf{0.987} & 0.124 & \textbf{0.749} & 0.457 & \textbf{0.989} & 0.178 & \textbf{0.748} & 0.987 & \underline{\textbf{1.000}} & 0.821 & \textbf{0.979} \\
Nomic-v2   & \underline{0.958} & \textbf{0.994} & \underline{0.600} & \textbf{0.765} & \underline{0.944} & \textbf{0.996} & \underline{0.596} & \textbf{0.767} & \underline{1.000} & \underline{1.000} & \underline{0.968} & \textbf{0.979} \\
\\
MXB-large  & 0.027 & \textbf{0.356} & 0.012 & \textbf{0.117} & 0.087 & \textbf{0.427} & 0.033 & \textbf{0.168} & 0.366 & \textbf{0.910} & 0.125 & \textbf{0.632} \\
GIST-large & 0.045 & \textbf{0.298} & 0.014 & \textbf{0.090} & 0.116 & \textbf{0.385} & 0.039 & \textbf{0.152} & 0.415 & \textbf{0.880} & 0.144 & \textbf{0.551} \\
E5-large   & 0.503 & \underline{\textbf{0.995}} & 0.175 & \underline{\textbf{0.826}} & 0.722 & \underline{\textbf{0.998}} & 0.300 & \underline{\textbf{0.852}} & 0.988 & \underline{\textbf{1.000}} & 0.831 & \underline{\textbf{0.983}} \\
\bottomrule
\end{tabular}
}
\caption{Document similarity search results on paired ``event-level'' multilingual Swiss Court Case Summaries (\ref{sec:setup:event}), before and after linear concept erasure. Here, the concept is the document's \emph{language}. Once again, erasure significantly improves recall of the paired item in all cases. The only exception is one instance where retrieval performance is already perfect before erasure. In some cases, erasure even allows smaller models to outperform their larger counterparts.}
\label{tab:event-scores-swiss}
\end{table*}

\subsubsection{Qualitative Analysis}
To investigate which semantic features contribute most and least to changes before and after LEACE, we conducted a qualitative analysis using pairs of legal case summaries sourced from Wikipedia and LexisNexis, employing multilingual E5 as the embedding model.

We observe that when baseline similarity is already high due to shared style or genre, LEACE encounters fewer residual confounders to remove, thus yielding relatively smaller improvements. Conversely, when baseline similarity is lower owing to clear stylistic or domain differences, LEACE proves particularly effective, as it targets and removes pronounced confounding signals, leading to greater gains. 

Our examination of legal case summaries provides insights into these dynamics. Summaries that experience the largest changes after applying LEACE are generally shorter, focus primarily on legal rulings, and lack factual idiosyncrasies. Without LEACE-based deconfounding, these summaries present substantial challenges for similarity-based text retrieval because of their limited semantic overlap regarding specific facts and rules (see also \citealt{fan2025lexam}). We provide examples in~\Cref{app:qa_1}.

On the other hand, summaries that display minimal change tend to incorporate both factual details and judgments, offer broader contextual framing, discuss subsequent impacts, and frequently integrate direct quotations from court decisions. An example can be found in~\Cref{app:qa_2}.

These findings underscore LEACE’s distinctive advantage in scenarios where domain experts can clearly identify and specify features irrelevant to downstream similarity tasks, highlighting its potential value within human-in-the-loop frameworks that leverage expert knowledge to detect and eliminate confounding factors. They also show LEACE is less useful when relevance relies on higher-order discourse features such as blending facts, judgments, and framing, which it cannot account for.

\section{Erasure helps, but can it hurt?}\label{sec:mteb}

The results from the previous section appear conclusive: linear concept erasure effectively removes spurious information from embeddings that distorts similarities.
At the same time, we must ask whether erasure might also degrade embeddings in subtle ways that our evaluations fail to detect.
Although LEACE is designed to minimize unwanted distortions, the trained eraser may inadvertently remove ``desirable'' information that may support other tasks.

In this section, we address this question through additional evaluations on out-of-distribution (OOD) benchmarks.
These experiments test whether applying an eraser trained for a specific domain unintentionally harms general-purpose semantic representations.
While our main experiments in the previous section focused on domain-specific differences, real-world deployment of embedding models often requires robust cross-domain performance.
We thus benchmark our models against diverse evaluation datasets from MTEB \cite{muennighoff-etal-2023-mteb} to assess whether erasers trained to isolate certain information also degrade performance in unrelated tasks.

\subsection{Data and Methods}

We focus on two sentence embedding models: MiniLM and E5-base-v2 \cite{wang2022text}.
Each model is paired with two trained concept erasers: the CAP eraser, trained to remove the source from the \textit{Bill--Newspapers} pair, and the Legal eraser, trained on \textit{LexisNexis--Wikipedia}.
This results in four models-eraser combinations per task.

We apply these combinations to retrieval and semantic texutal similarity (STS) tasks from MTEB \cite{muennighoff-etal-2023-mteb}: (1) Legal Retrieval tasks, (2) News Retrieval tasks \cite{thakur2021beir}, and (3) STS News tasks.
These benchmarks differ in domain, structure, and evaluation metrics, offering a comprehensive perspective on erased embedding behavior in out-of-domain settings.
For each benchmark, we compare the performance of the original model embeddings to the same embeddings after applying the trained LEACE erasers.%
\begin{figure}[ht]
    \centering
    \includegraphics[width=1.0\linewidth]{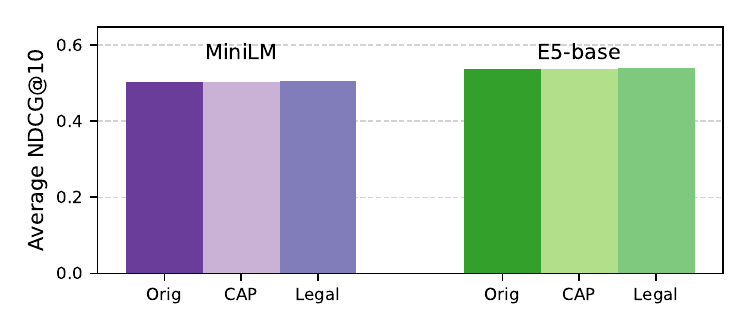}
    \caption{An eraser trained on embeddings from one dataset does not degrade embeddings from a different dataset. Erasers fit to the CAP and SCOTUS data (\S \ref{sec:setup}) are applied to embeddings (MiniLM and E5-base-v2) from five legal retrieval tasks. Each triplet of same-color bars compares the average NDCG@10 for the base and erased embeddings.}
    \label{fig:legal_retrieval}
\end{figure}
\subsection{MTEB Results}
We report a selection of results here, again emphasizing that our hope is not to improve benchmark results, but to \emph{avoid making them worse} (full results in Appendix~\ref{app:mteb}).

\paragraph{Retrieval.} On both the legal and news retrieval tasks, the trained erasers do not harm performance (as measured by the average NDCG@10). See \cref{fig:legal_retrieval} for legal retrieval; per-task performance (\cref{fig:lr-mteb}) and news retrieval (\cref{fig:news_retrieval}) are in the appendix.
Given the domain overlap, we had hypothesized that the Legal eraser might improve legal retrieval somewhat, but only one task sees a marginal improvement (AILACasedocs), from 0.197 to 0.218 (\cref{tab:lr-mteb} in appendix).
That said, the results are still positive overall, indicating that both the CAP and Legal erasers operate robustly in OOD retrieval tasks with both small and large models.

\paragraph{Semantic Textual Similarity.} We evaluate the four model–eraser combinations on eight well-established STS benchmarks, covering both monolingual and crosslingual settings in the news domain (\cref{fig:sts_news} and \cref{tab:sts-mteb} in the appendix).
The evaluation metric is the Spearman correlation between embedding cosine similarities and ground-truth semantic similarity.
LEACE does not degrade performance over tasks, with most scores either unchanged or showing negligible increases.

Across the retrieval and semantic similarity evaluations, LEACE consistently preserves the quality of the embedding space while effectively removing targeted conceptual signals.
These results reinforce its utility as a lightweight and reliable method for concept erasure.

\begin{figure}[t]
    \centering
    \includegraphics[width=.9\linewidth]{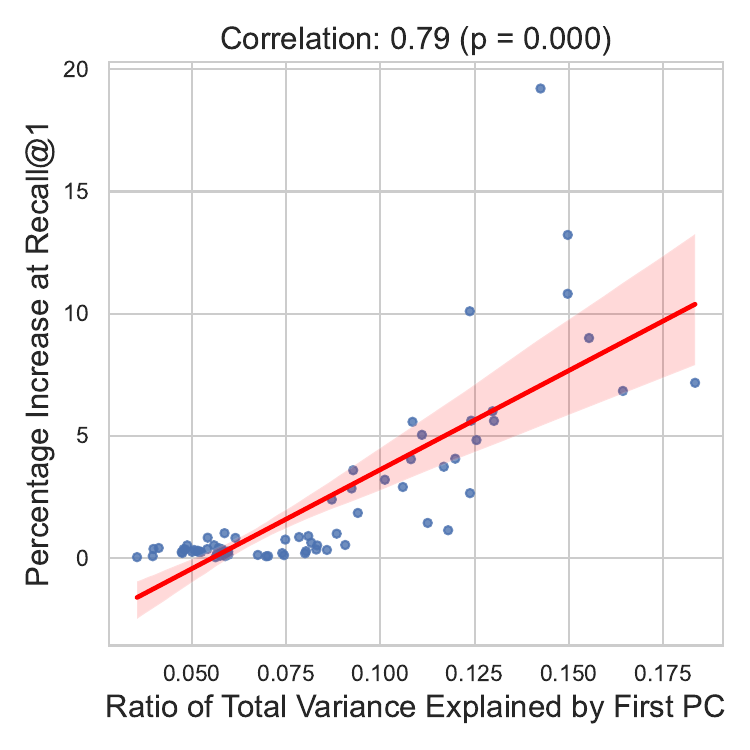}
    \caption{Relationship between variance explained by PC1 in the original embeddings and Recall@1 improvement after LEACE. Each point corresponds to a dataset setting in the event-level evaluation.}
    \label{fig:pca_ratio}
\end{figure}

\section{Additional Findings}

\paragraph{Relating LEACE to PCA} Why does LEACE work in these settings?
Here, we consider its relationship to Principal Component Analysis (PCA).

Taking the embeddings of the German-French Swiss court summaries, the first principal component (PC1) forms two clearly separable clusters corresponding directly to the text's language (\cref{fig:erasure-illustration}).
After applying LEACE, the clusters collapse into a single, overlapping distribution, an indication that language identity is no longer linearly separable in the embedding space.

To better understand when LEACE is effective, we investigate how the structural characteristics of the original embedding space relate to observed performance improvements. 
Specifically, we hypothesize that LEACE provides greater performance gains when the removable concept is prominently encoded within the embedding space.

To test this, we apply PCA to the original embeddings from each event-level dataset (SCOTUS, SemEval, Swiss Court Cases) and record the proportion of total variance explained by PC1.
A high proportion of explained variance suggests that PC1 encodes a dominant direction in the embedding space, which will tend to correspond to the concept targeted by LEACE (i.e., the source or language, per \cref{fig:erasure-illustration}).
Indeed, \cref{fig:pca_ratio} shows a strong positive correlation ($r=0.79$, $p<0.001$) between the proportion of variance explained by PC1 and the percentage improvement in Recall@1.
This result indicates that LEACE is more effective when the removable concept aligns with dominant directions in the embedding space. %

Given these findings, one might ask why not use PCA for erasure instead, following \citet{Bolukbasi2016ManITA}. 
We observe positive but less consistent results than LEACE on our tasks, along with a strong degradation in MTEB performance (\cref{app:baseline}).

\paragraph{A new task: bitext mining} Erasure improves already multilingual models on with multilingual tasks, so can it help with \emph{bitext mining}---retrieving translation pairs via similarity search?
Our further experiments show that improvements are not uniformly strong, but we do achieve state-of-the-art results on a few leaderboard tasks from  \citet{enevoldsen2025mmtebmassivemultilingualtext}, and erasure never reduces performance (details in~\cref{app:bitext}).

\section{Conclusion}
For applied practitioners working with large text collections from multiple sources or languages---a common scenario---our results offer a clear recommendation: apply linear erasure to document embeddings before use to remove confounding information. While there may be cases where it is less effective, the method does not appear to harm representations (see below) and incurs only minimal computational cost.

\section*{Limitations}

The primary limitation of our method is its dependence on per-document metadata or labels. 
If an undesirable low-level pattern in the data distribution is suspected but not known---say, an unreported change in how a corpus was collected over a long time period---then the user must first apply some possibly-unsupervised labeling method. 
Although confounder labels are available for many tasks, reliance on such labels constrains the broader applicability of our proposed methods. 
We explored approaches that automatically generate features; however, these did not lead to measurable improvements in downstream retrieval or similarity comparison tasks. 
Other works have developed unsupervised techniques to debias or erase neural representations \cite{kim2019learning, seo2022unsupervised, yangscissor}. We leave a deeper exploration of such methods in our context for future work.
One direction is to integrate sparse autoencoders \cite[e.g.,][]{movvasparse, pauloautomatically} that extract features whose utility can be interpreted and validated by human experts.

Another shortcoming arises when metadata is available but the categories are too numerous relative to the total number of items.
For example, in the paired \emph{within}-language (en--en) SemEval Task 8 news articles, the data originate from dozens of sources, many of which are represented by only a handful of articles.
In contrast to removing the language label in the multilingual data (\cref{tab:model-scores-pert-semeval}), removing the source label does not improve retrieval results over the baseline.
A possible direction for future work is to first merge similar sources into broader categories (e.g., local vs. national newspapers) before applying label erasure.

A final limitation was first noted by \citet{huang-etal-2024-language}, who used LEACE as a baseline in multilingual retrieval contexts. 
While they similarly removed language information, their results were mixed, suggesting that LEACE may not be effective in all settings. 
One hypothesis is that our tasks, though realistic, differ from the standard benchmark data on which models are typically trained, leading to saturated in-domain performance that does not transfer well out-of-domain. 
Another possibility is that retrieval setups---with their distinct (short query, document) structure, as opposed to our (document, document) structure---may be less amenable to erasure. We plan to explore these hypotheses to help explain such discrepancies.

\section*{Acknowledgments} Funding provided in part by the ETH AI Center and Innosuisse (project number: 122.540 IP-ICT). Thanks to Maria Antoniak for pointing out the connection to Authorless Topic Models, and to Jacob Conway for his insightful suggestions on task transferability.

\bibliography{custom,anthology_post2020,anthology_pre2020}

\appendix

\clearpage
\section{Bitext Mining Results}\label{app:bitext}
Our gains on multilingual tasks with already multilingual embeddings motivate us to ask whether erasure can benefit already ``saturated'' leaderboard tasks that cover multiple languages.
To this end, we focus on \emph{bitext mining}: given pairs of sentences in different languages, the goal is to retrieve a specific sentence in the target language given a ``query'' sentence in the source language (typically a translation;  $F_1$ is the standard metric).
We collect all 28 tasks available through the MTEB package at the time of writing \cite{muennighoff-etal-2023-mteb} and use \texttt{E5-large-instruct}, one of the best-performing models on the leaderboard.

In several cases, there is a marked increase, yielding state-of-the-art scores on three tasks that appear on the public leaderboard (even with a different base model class,  \cref{tab:bitext_results} in appendix).
Generally, though, the improvements are much smaller than those in our main experiments, with over half of the 28 tasks showing 
 less than a 0.01 change (although no tasks decrease more than $-0.01$).
First applying LEACE is therefore a simple step when bitext mining; even if it may not always help, it is unlikely to hurt.

\begin{table}[h!tbp]
\small
\begin{tabular}{lrrr}
\toprule
 & \multicolumn{2}{c}{$F_1$} & \\
 \cmidrule{2-3}
 & Before & After & $\Delta$ \\
\midrule
SynPerChatbotSumS & 0.283 & 0.500 & 0.217 \\
SAMSumFa & 0.811 & 0.943 & 0.132 \\
SynPerChatbotRAGSumS & 0.560 & 0.680 & 0.120 \\
RomaTales & 0.201 & 0.263 & 0.062 \\
SRNCorpus & 0.500 & 0.551 & 0.051 \\
NusaX* & 0.853 & \textbf{0.892} & 0.039 \\
NollySenti* & 0.807 & \textbf{0.839} & 0.032 \\
NusaTranslation* & 0.851 & \textbf{0.876} & 0.025 \\ 
LinceMT & 0.487 & 0.506 & 0.019 \\
Bornholm* & 0.560 & 0.578 & 0.018 \\
IN22Conv & 0.626 & 0.637 & 0.011 \\
Phinc & 0.855 & 0.867 & 0.011 \\
\\
\multicolumn{3}{l}{\emph{Number of tasks with} $|\Delta|<0.01$} & \emph{15} \\
 \bottomrule
\end{tabular}
\caption{$F_1$ on MTEB Bitext Mining Tasks before and after erasing the language ID, for \texttt{E5-large-instruct}. Gains are substantial in a few cases, sometimes improving over the reported state-of-the-art on MTEB (tasks with * appear on the public leaderboard, improvements over SotA in \textbf{bold}).}\label{tab:bitext_results}
\end{table}

\section{SemEval English \& Non-English News Results}

The results on testing LEACE on the SemEval 2022 Task 8 dataset are presented in \cref{tab:model-scores-pert-semeval}. All models benefit from LEACE, with consistent improvements in both Recall@10 and Recall@1. The E5-small model shows the strongest gains overall: +0.202 (Recall@10) and +0.236 (Recall@1). High-performing large models like E5-large and MXB-large achieve further enhancements of up to +0.156 in Recall@1. Smaller models also gain notable increases. For instance, MiniLM gains +0.183 (Recall@10) and +0.127 (Recall@1), respectively. These improvements highlight LEACE’s utility in reducing source bias and improving semantic alignment in multilingual event representations. Nomic-v2, which already has high scores before LEACE, showed modest increases, likely due to saturation. In general, LEACE proves effective even under high-resource, multilingual scenarios.

\begin{table}[ht]
\centering
\small
\begin{tabular}{l*{2}{c}*{2}{c}}
\toprule
& \multicolumn{2}{c}{\texttt{Recall@10}} & \multicolumn{2}{c}{\texttt{Recall@1}} \\
\cmidrule(lr){2-3} \cmidrule(lr){4-5} 
Model & Before & After & Before & After \\
\midrule
MiniLM     & 0.350 & \textbf{0.533} & 0.150 & \textbf{0.277} \\
GIST-small & 0.497 & \textbf{0.636} & 0.247 & \textbf{0.372} \\
E5-small   & 0.614 & \textbf{0.816} & 0.318 & \textbf{0.554} \\
\\ 
MPNet      & 0.557 & \textbf{0.664} & 0.262 & \textbf{0.347} \\
GIST-base  & 0.564 & \textbf{0.694} & 0.301 & \textbf{0.402} \\
E5-base    & 0.777 & \textbf{0.859} & 0.466 & \textbf{0.601} \\
Nomic-v2   & \underline{0.892} & \underline{\textbf{0.906}} & \underline{0.637} & \underline{\textbf{0.651}} \\
\\ 
MXB-large  & 0.527 & \textbf{0.691} & 0.250 & \textbf{0.390} \\
GIST-large & 0.624 & \textbf{0.734} & 0.332 & \textbf{0.428} \\
E5-large   & 0.747 & \textbf{0.866} & 0.436 & \textbf{0.592} \\
\bottomrule
\end{tabular}
\caption{Results on SemEval English \& Non-English News Articles}
\label{tab:model-scores-pert-semeval}
\end{table}

\section{MTEB Evaluation Results}
\label{app:mteb}

\begin{figure*}[ht]
    \centering
    \includegraphics[width=\linewidth]{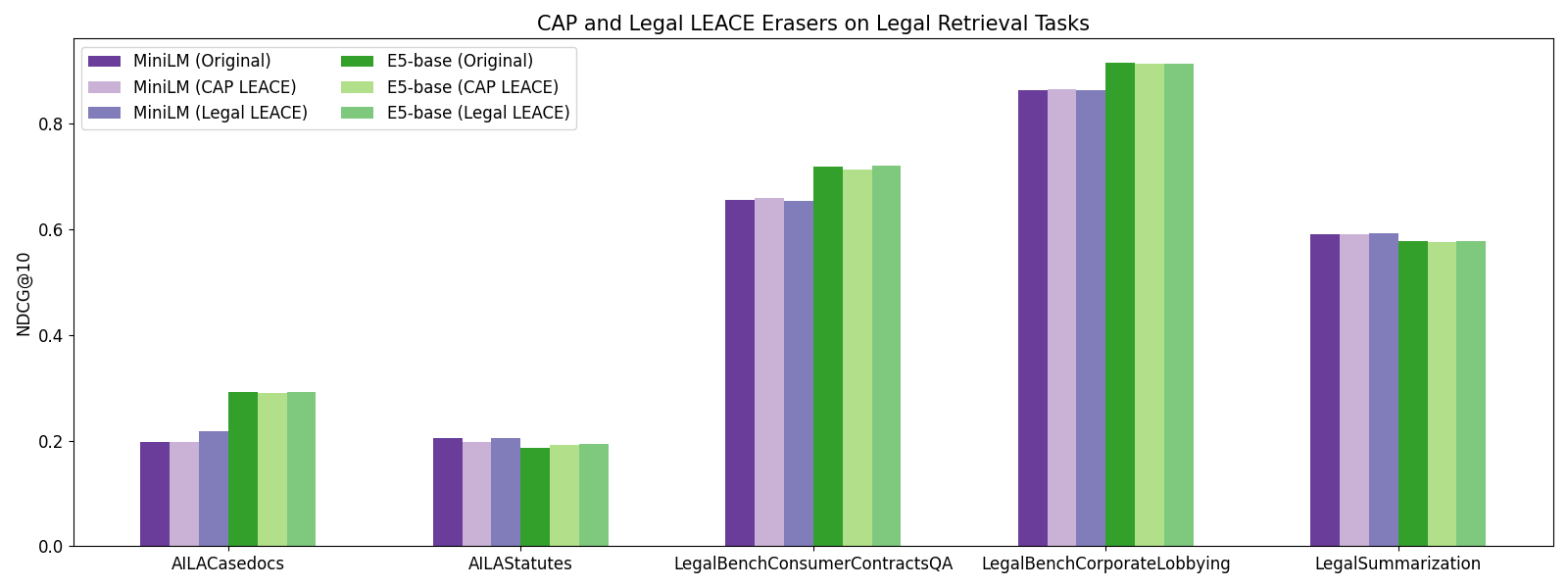}
    \caption{Performance of CAP and Legal erasers across three news retrieval tasks. Each group of bars compares the base and LEACE-erased models for MiniLM and E5-base-v2 embeddings.}
    \label{fig:lr-mteb}
\end{figure*}

\begin{figure*}[ht]
    \centering
    \includegraphics[width=\linewidth]{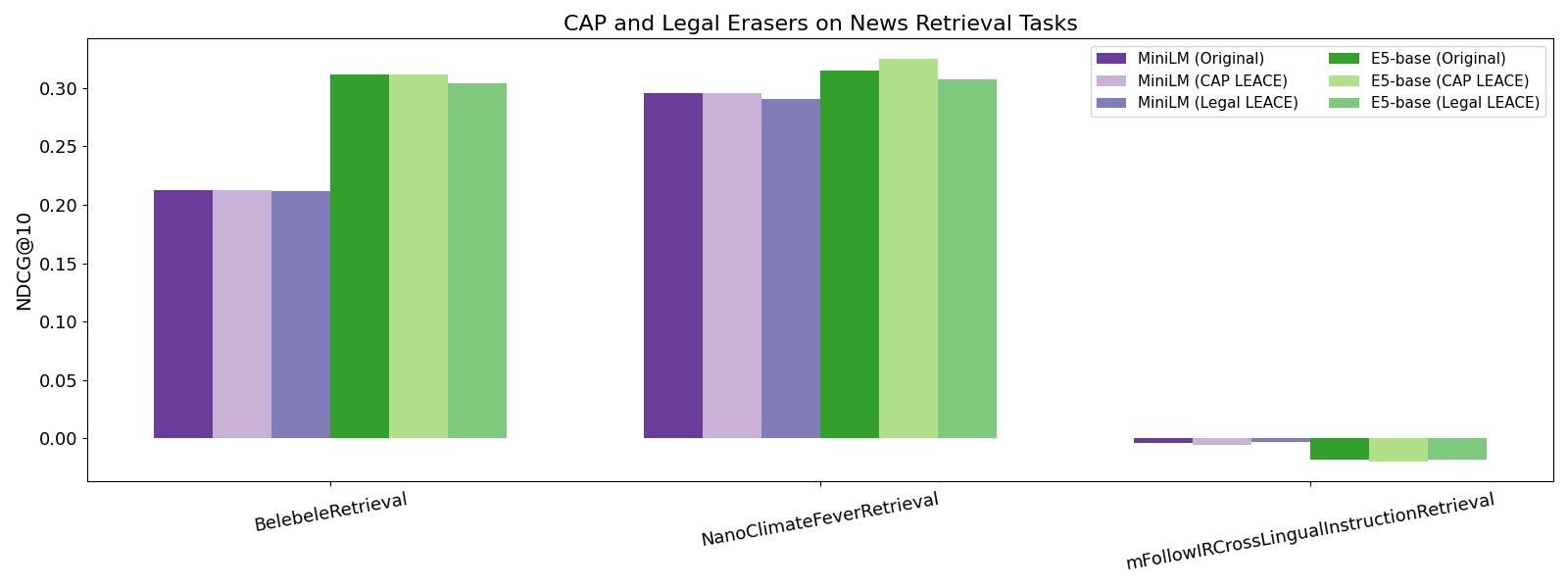}
    \caption{Performance of CAP and Legal erasers across three news retrieval tasks. Each group of bars compares the base and LEACE-erased models for MiniLM and E5-base-v2 embeddings.}
    \label{fig:news_retrieval}
\end{figure*}

\begin{figure*}[ht]
    \centering
    \includegraphics[width=\linewidth]{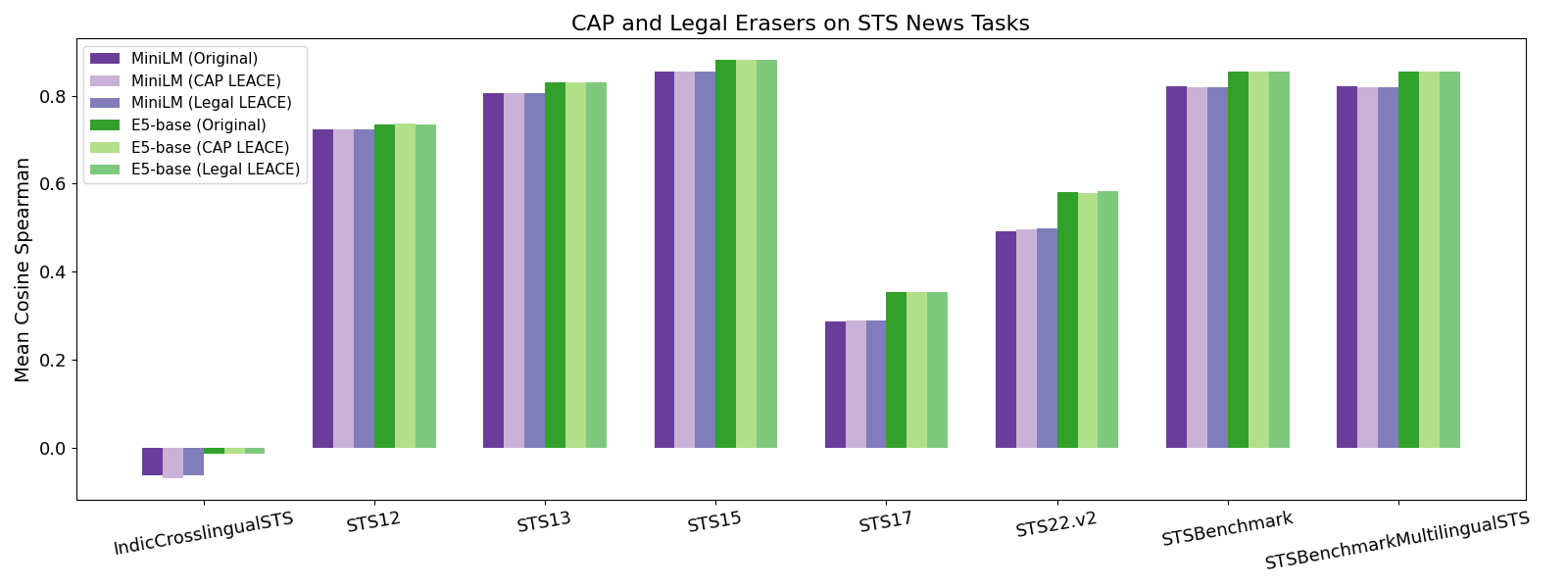}
    \caption{Performance of CAP and Legal erasers across eight STS news tasks. Each group of bars compares the base and LEACE-erased models for MiniLM and E5-base-v2 embeddings. }
    \label{fig:sts_news}
\end{figure*}

We report the full evaluation results of the CAP and Legal erasers on three MTEB benchmark groups: Legal Retrieval, News Retrieval, and STS News Tasks. Each setting involves comparing model performance before and after LEACE-based erasure, across two embedding models (MiniLM and E5-base), as shown in \cref{tab:lr-mteb}, \cref{tab:news-mteb}, and \cref{tab:sts-mteb} and \cref{fig:lr-mteb}, \cref{fig:news_retrieval} and \cref{fig:sts_news}.

\begin{table*}[ht]
\centering
\setlength{\tabcolsep}{4pt}        %
\renewcommand{\arraystretch}{1.1}  %
\footnotesize                      %
\resizebox{\textwidth}{!}{         %
\begin{tabular}{lcccccc}
\toprule
\multirow{2}{*}{\textbf{Task}} 
& \multicolumn{3}{c}{MiniLM} 
& \multicolumn{3}{c}{E5-base} \\
\cmidrule(lr){2-4} \cmidrule(lr){5-7}
 & Before & After (CAP) & After (Legal) 
 & Before & After (CAP) & After (Legal) \\
\midrule
\texttt{AILACasedocs}        & 0.197 & 0.197 & 0.218 & 0.292 & 0.290 & 0.292 \\
\texttt{AILAStatutes}        & 0.205 & 0.196 & 0.205 & 0.186 & 0.191 & 0.193 \\
\texttt{ConsumerContractsQA} & 0.656 & 0.659 & 0.654 & 0.720 & 0.712 & 0.720 \\
\texttt{CorporateLobbying}   & 0.864 & 0.865 & 0.863 & 0.915 & 0.914 & 0.913 \\
\texttt{LegalSummarization}  & 0.590 & 0.591 & 0.592 & 0.577 & 0.576 & 0.578 \\
\bottomrule
\end{tabular}
}
\caption{Legal Retrieval Results on MTEB evaluated using NDCG@10. Each model (MiniLM, E5-base-v2) is tested with and without LEACE erasure, using both CAP and Legal erasers.}
\label{tab:lr-mteb}
\end{table*}

\begin{table*}[ht]
\small
\centering
\begin{tabular}{lcccccc}
\toprule
\multirow{2}{*}{\textbf{Task}} & \multicolumn{3}{c}{MiniLM} & \multicolumn{3}{c}{E5-base} \\
\cmidrule(lr){2-4} \cmidrule(lr){5-7}
 & Before & After (CAP) & After (Legal) & Before & After (CAP) & After (Legal) \\
\midrule
\texttt{BelebeleRetrieval}           & 0.212 & 0.212 & 0.211 & 0.312 & 0.311 & 0.303 \\
\texttt{NanoClimateFeverRetrieval}  & 0.296 & 0.296 & 0.291 & 0.315 & 0.325 & 0.307 \\
\texttt{mFollowIR (CrossLingual)}   & -0.004 & -0.005 & -0.003 & -0.018 & -0.019 & -0.018 \\
\bottomrule
\end{tabular}
\caption{News Retrieval Results on MTEB evaluated using NDCG@10. Each model (MiniLM, E5-base-v2) is evaluated before and after applying LEACE, using both CAP and Legal erasers.}
\label{tab:news-mteb}
\end{table*}

\begin{table*}[ht]
\small
\centering
\begin{tabular}{lcccccc}
\toprule
\multirow{2}{*}{\textbf{Task}} & \multicolumn{3}{c}{MiniLM} & \multicolumn{3}{c}{E5-base} \\
\cmidrule(lr){2-4} \cmidrule(lr){5-7}
 & Before & After (CAP) & After (Legal) & Before & After (CAP) & After (Legal) \\
\midrule
\texttt{IndicCrosslingualSTS}         & -0.063 & -0.070 & -0.062 & -0.013 & -0.012 & -0.013 \\
\texttt{STS12}                        & 0.724  & 0.724  & 0.723  & 0.735  & 0.736  & 0.735 \\
\texttt{STS13}                        & 0.806  & 0.806  & 0.806  & 0.830  & 0.830  & 0.830 \\
\texttt{STS15}                        & 0.854  & 0.854  & 0.854  & 0.882  & 0.882  & 0.882 \\
\texttt{STS17}                        & 0.288  & 0.289  & 0.288  & 0.354  & 0.355  & 0.353 \\
\texttt{STS22.v2}                     & 0.492  & 0.496  & 0.499  & 0.581  & 0.578  & 0.583 \\
\texttt{STSBenchmark}                 & 0.820  & 0.820  & 0.820  & 0.855  & 0.855  & 0.855 \\
\texttt{STSBenchmarkMultilingualSTS}  & 0.820  & 0.820  & 0.820  & 0.855  & 0.855  & 0.855 \\
\bottomrule
\end{tabular}
\caption{STS News Results on MTEB evaluated using the mean cosine Spearman score. Each model (MiniLM, E5-base-v2) is evaluated before and after LEACE, using both CAP and Legal erasers.}
\label{tab:sts-mteb}
\end{table*}

\section{Sources of MTEB Tasks}

We list below the original sources for the datasets used from the MTEB benchmark \cite{muennighoff-etal-2023-mteb, enevoldsen2025mmtebmassivemultilingualtext}:

\begin{itemize}
	\item Legal retrieval tasks: \textbf{AILACasedocs} and \textbf{AILAStatutes} \cite{paheli_bhattacharya_2020_4063986}, \textbf{LegalBenchConsumerContractsQA} \cite{wang2025acordexpertannotatedretrievaldataset, koreeda-manning-2021-contractnli-dataset}, \textbf{LegalBenchCorporateLobbying} \cite{guha2023legalbench, holzenberger-van-durme-2021-factoring, lippi2019claudette, ravichander-etal-2019-question, wang-etal-2023-maud, wilson-etal-2016-creation, zheng2021does, zimmeck2019maps}, \textbf{LegalSummarization} \cite{manor-li-2019-plain}.
    
	\item News retrieval tasks: \textbf{BelebeleRetrieval} \cite{bandarkar-etal-2024-belebele},  \textbf{NanoClimateFeverRetrieval} \cite{diggelmann2021climatefever}, \textbf{mFollowIRCrossLingualInstructionRetrieval} \cite{weller2025mfollowir}.
    
    \item STS news tasks: \textbf{IndicCrosslingualSTS} \cite{ramesh-etal-2022-samanantar}, \textbf{STS12} \cite{agirre-etal-2012-semeval}, \textbf{STS13} \cite{agirre-etal-2013-sem}, \textbf{STS15} \cite{bicici-2015-rtm}, \textbf{STS17} \cite{cer-etal-2017-semeval},  \textbf{STS22} \cite{chen-etal-2022-semeval}, \textbf{STSBenchmark} and \textbf{STSBenchmarkMultilingualSTS} \cite{huggingface:dataset:stsb_multi_mt}.
    
\end{itemize}

\section{Additional PCA Analysis}\label{app:baseline}
We create a baseline by removing PC1 from the embedding space, and evaluate it in the event-level setting using the SCOTUS dataset (\cref{tab:event-scores-scotus}).
Overall, the baseline occasionally helps and can even marginally outperform LEACE in a few cases, but its effectiveness appears unstable, heavily dependent on the particular setting and model used (although it is effective for the E5 family for most configurations). 
Furthermore, in some cases, it performs worse than applying no erasure at all.
There is also a final catch: removing the learned PC1 from OOD embeddings \emph{does} dramatically degrade performance on MTEB tasks (\cref{tab:lr-pc1}), unlike LEACE (\cref{fig:legal_retrieval_comparison}).

\subsection{Event-Level Results on SCOTUS Case Summaries}

\cref{tab:event-scores-scotus} reveals the results of applying the baseline, which removes the first principal component (PC1) from the embedding space, in the event-level setting on the SCOTUS dataset. While it sometimes improves over the original embeddings and occasionally outperforms LEACE (especially for the E5 family), its performance is inconsistent across models and configurations, and it can underperform even relative to no erasure.
 
\begin{table*}[ht]
\centering
\resizebox{\textwidth}{!}{%
\begin{tabular}{l
    *{3}{cc}  %
    *{3}{cc}  %
    *{3}{cc}  %
}
\toprule
\multirow{3}{*}{Model}
& \multicolumn{6}{c}{LexisNexis \& Wikipedia}
& \multicolumn{6}{c}{LexisNexis \& Oyez}
& \multicolumn{6}{c}{Oyez \& Wikipedia} \\
\cmidrule(lr){2-7} \cmidrule(lr){8-13} \cmidrule(lr){14-19}
& \multicolumn{3}{c}{\texttt{Recall@10}} & \multicolumn{3}{c}{\texttt{Recall@1}}
& \multicolumn{3}{c}{\texttt{Recall@10}} & \multicolumn{3}{c}{\texttt{Recall@1}}
& \multicolumn{3}{c}{\texttt{Recall@10}} & \multicolumn{3}{c}{\texttt{Recall@1}} \\
\cmidrule(lr){2-4} \cmidrule(lr){5-7}
\cmidrule(lr){8-10} \cmidrule(lr){11-13}
\cmidrule(lr){14-16} \cmidrule(lr){17-19}
& Before & After & Baseline & Before & After & Baseline
& Before & After & Baseline & Before & After & Baseline
& Before & After & Baseline & Before & After & Baseline \\
\midrule
MiniLM     & 
0.487 & \textbf{0.606} & 0.478 & 0.231 & \textbf{0.313} & 0.231 & 
0.890 & \textbf{0.899} & 0.883 & 0.651 & \textbf{0.693} & 0.646 & 
0.850 & \textbf{0.924} & 0.869 & 0.623 & \textbf{0.747} & 0.670 \\
GIST-small & 
0.563 & \textbf{0.656} & 0.547 & 0.261 & \textbf{0.325} & 0.254 & 
0.918 & \textbf{0.943} & 0.920 & 0.702 & \textbf{0.778} & 0.701 & 
0.762 & \textbf{0.844} & 0.776 & 0.478 & \textbf{0.599} & 0.500 \\
E5-small   & 
0.421 & 0.673 & \textbf{0.675} & 0.176 & 0.353 & \textbf{0.356} & 
0.830 & \textbf{0.939} & \textbf{0.939} & 0.563 & \textbf{0.789} & \textbf{0.789} & 
0.689 & \textbf{0.951} & 0.950 & 0.398 & 0.752 & \textbf{0.753} \\
\\
MPNet      & 
0.566 & \textbf{0.666} & 0.552 & 0.259 & \textbf{0.337} & 0.257 & 
0.926 & \textbf{0.943} & 0.925 & 0.724 & \textbf{0.775} & 0.722 &
0.856 & \textbf{0.911} & 0.862 & 0.565 & \textbf{0.678} & 0.574 \\
GIST-base  & 
0.646 & \textbf{0.757} & 0.636 & \underline{0.308} & \textbf{0.412} & 0.309 & 
0.939 & \textbf{0.963} & 0.936 & 0.727 & \textbf{0.819} & 0.725 &
0.880 & \textbf{0.950} & 0.917 & 0.628 & \textbf{0.773} & 0.701\\
E5-base    & 
0.414 & \textbf{0.660} & \textbf{0.660} & 0.188 & 0.341 & \textbf{0.344} & 
0.830 & \textbf{0.940} & 0.939 & 0.575 & \textbf{0.758} & 0.755 &
0.650 & \textbf{0.942} & \textbf{0.942} & 0.371 & 0.737 & \textbf{0.738} \\
Nomic-v2   & 
0.530 & 0.701 & \textbf{0.703} & 0.254 & \textbf{0.384} & 0.382 & 
\underline{0.950} & \textbf{0.966} & 0.948 & 0.770 & \textbf{0.820} & 0.767 & 
0.903 & \underline{0.978} & \underline{\textbf{0.981}} & 0.658 & \textbf{0.819} & \underline{0.818} \\
\\
MXB-large  & 
0.537 & \textbf{0.703} & 0.627 & 0.249 & \textbf{0.376} & 0.321 & 
0.928 & \textbf{0.958} & 0.933 & 0.720 & \textbf{0.805} & 0.729 & 
0.883 & \textbf{0.960} & 0.919 & 0.654 & \textbf{0.819} & 0.737 \\
GIST-large & 
\underline{0.657} & \underline{\textbf{0.770}} & 0.641 & 0.305 & \underline{\textbf{0.414}} & 0.300 & 
\underline{0.954} & \textbf{\underline{0.967}} & \underline{0.954} & \underline{0.787} & \underline{\textbf{0.834}} & 0.787 &
0.947 & \underline{\textbf{0.971}} & 0.944 & \underline{0.760} & \underline{\textbf{0.826}} & 0.772 \\
E5-large   & 
0.479 & \textbf{0.720} & \underline{0.717} & 0.209 & 0.381 & \textbf{\underline{0.388}} & 
0.864 & \textbf{0.949} & \textbf{0.949} & 0.636 & \textbf{0.791} & \underline{0.790} & 
0.765 & \textbf{0.964} & 0.963 & 0.489 & \textbf{0.792} & \textbf{0.792} \\
\bottomrule
\end{tabular}
}
\caption{Event-Level Results on SCOTUS Case Summaries}
\label{tab:event-scores-scotus}
\end{table*}

\subsection{MTEB Evaluation Results}

\begin{figure*}[ht]
    \centering
    \includegraphics[width=\linewidth]{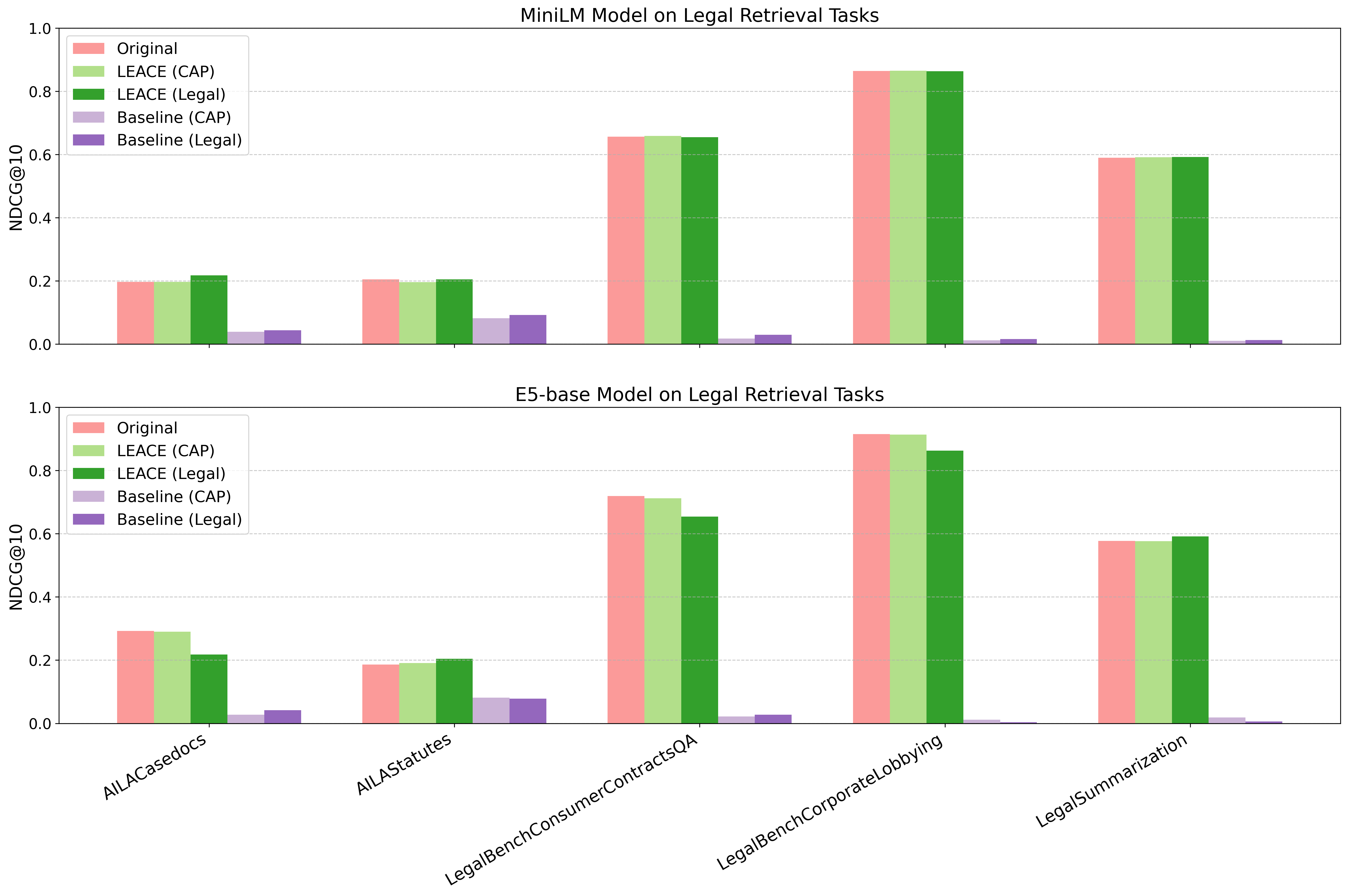}
    \caption{Comparison of average NDCG@10 scores across five MTEB legal retrieval tasks. Each group of bars compares the original, LEACE-erased and baseline models for MiniLM and E5-base-v2 models.}
    \label{fig:legal_retrieval_comparison}
\end{figure*}

\cref{tab:lr-pc1} shows the results of applying the baselines, derived from both CAP and SCOTUS datasets, on the MTEB legal retrieval tasks. In all cases, this PC1 removal leads to a drastic performance drop for both MiniLM and E5-base models. As observed in the comparison between the two approaches in \cref{fig:legal_retrieval_comparison}, in contrast, LEACE erasures maintain retrieval quality, highlighting its robustness.

\begin{table*}[ht]
\centering
\begin{tabular}{lccc ccc}
\toprule
\multirow{2}{*}{\textbf{Task}} 
& \multicolumn{3}{c}{MiniLM} 
& \multicolumn{3}{c}{E5-base} \\
\cmidrule(lr){2-4} \cmidrule(lr){5-7}
 & Before & After (CAP) & After (Legal) 
 & Before & After (CAP) & After (Legal) \\
\midrule
\texttt{AILACasedocs}                    & 0.197 & 0.039 & 0.044 & 0.292 & 0.027 & 0.042 \\
\texttt{AILAStatutes}                    & 0.205 & 0.082 & 0.092 & 0.186 & 0.081 & 0.079 \\
\texttt{ContractsQA}  & 0.656 & 0.018 & 0.029 & 0.720 & 0.022 & 0.028 \\
\texttt{CorporateLobbying}    & 0.864 & 0.012 & 0.016 & 0.915 & 0.012 & 0.004 \\
\texttt{LegalSummarization}             & 0.590 & 0.011 & 0.012 & 0.577 & 0.018 & 0.006 \\
\bottomrule
\end{tabular}
\caption{Legal Retrieval Results on MTEB evaluated using NDCG@10. Each mode (MiniLM, E5-base-v2) is evaluated before and after applying baseline model (PC1 removal), using both CAP and Legal erasers.}
\label{tab:lr-pc1}
\end{table*}

\begin{figure}[ht]
    \centering
    \includegraphics[width=\linewidth]{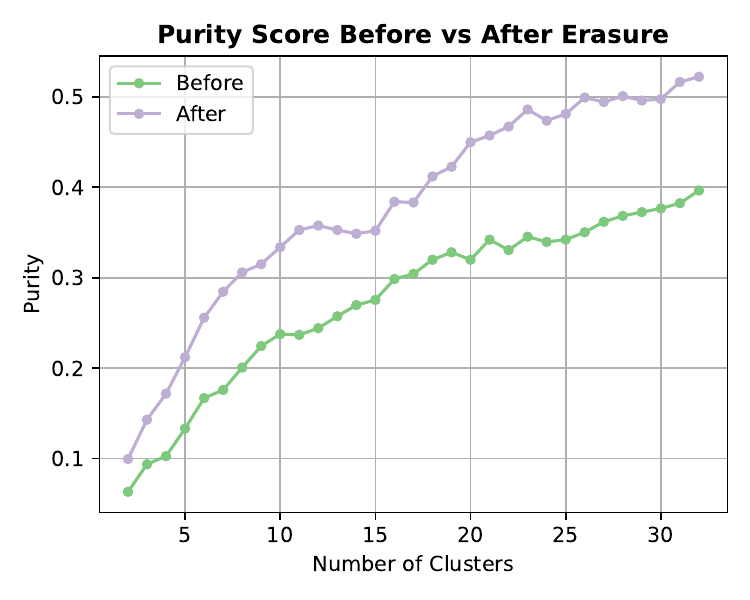}
    \caption{Purity score before vs. after LEACE erasure under different cluster counts, using data from CAP news articles and congressional bills.}
    \label{fig:purity}
\end{figure}

\section{Examples for Qualitative Analysis}
\label{app:qa}

We provide here examples for legal case summaries that experience the largest and least changes after applying LEACE.

\subsection{Examples with The Largest Changes}
\label{app:qa_1}

\begin{itemize}
	\item \textit{Riggins v. Nevada, 504 U.S. 127 (1992), is a U.S. Supreme Court case in which the court decided whether a mentally ill person can be forced to take antipsychotic medication  while they are on trial to allow the state to make sure they remain competent during the trial.}
    \item \textit{Benton v. Maryland, 395 U.S. 784 (1969), is a Supreme Court of the United States decision concerning double jeopardy. Benton ruled that the Double Jeopardy Clause of the Fifth Amendment applies to the states. In doing so, Benton expressly overruled Palko v. Connecticut.}
    \item \textit{Rutan v. Republican Party of Illinois, 497 U.S. 62 (1990), was a United States Supreme Court decision that held that the First Amendment forbids a government entity from basing its decision to promote, transfer, recall, or hire low-level public employees based upon their party affiliation.}    
\end{itemize}

\subsection{An Example with The Least Changes}
\label{app:qa_2}

\begin{itemize}
	\item \textit{Cuomo v. Clearing House Association, L.L.C., 557 U.S. 519 (2009), was a case decided by the United States Supreme Court. In a 5–4 decision, the court determined that a federal banking regulation did not pre-empt the ability of states to enforce their own fair-lending laws. The Court determined that the Office of the Comptroller of the Currency is the sole regulator of national banks but it does not have the authority under the National Bank Act to pre-empt state law enforcement against national banks. The case came out of an interpretation of the US Treasury Department's Office of the Comptroller of the Currency which had blocked an investigation by New York into lending practices. The OCC claimed that the 1864 National Bank Act bars states from enforcing their own laws against national banks. Justice Scalia stated in the opinion that while the OCC has "visitorial powers," the right to examine the affairs of a corporation, that does not mean that it has the exclusive right to enforcement. "A sovereign's 'visitorial powers' and its power to enforce the law are two different things. Contrary to what the [OCC's] regulation says, the National Bank Act pre-empts only the former." Scalia noted that states "have always enforced their general laws against national banks—and have enforced their banking-related laws against national banks for at least 85 years." The case is notable for the justices composing the 5-4 majority, which included the liberal justices (John Paul Stevens, David Souter, Ruth Bader Ginsburg, and Stephen Breyer) along with the conservative Scalia, who authored the opinion. Justice Clarence Thomas, joined by Justices Samuel Alito, Anthony Kennedy, and Chief Justice John Roberts, wrote a dissent. The case is further notable for the suggested relationship of this OCC decision to the 2008 financial crisis.}    
\end{itemize}

\section{Embedding model information} 
We list characteristics of the embedding models in \cref{tab:embed}.

\begin{table}[ht]
\centering
\small
\begin{tabular}{l*{2}{c}*{2}{c}}
\toprule
Models & \#Dims & \#Params & Multilingual & IFT \\
\midrule
MiniLM     & 384  & 22.7M  & & \\
GIST-small & 384  & 33.4M  & & \\
E5-small   & 384  & 118M   & \checkmark&  \\
\\
MPNet      & 768  & 109M   & &  \\
GIST-base  & 768  & 109M   & &  \\
E5-base    & 768  & 278M   & \checkmark &  \\
Nomic-v2   & 768  & 475M   & \checkmark & \checkmark \\
\\
MXB-large  & 1,024 & 335M   & & \checkmark \\
GIST-large & 1,024 & 335M   & &  \\
E5-large   & 1,024 & 560M   & \checkmark & \\
\bottomrule
\end{tabular}
\caption{Embedding Models. We examine mono- and multilingual models spanning multiple parameter sizes and embedding dimensions.}
\label{tab:embed}
\end{table}

\section{Use of AI Assistants} 
We used AI assistants, including ChatGPT and Claude, for editing (e.g.\ grammar, spelling, and word choice), debugging code, and visualizing results.

\end{document}